\documentclass{article}

\usepackage{PRIMEarxiv}
\usepackage{amsmath}
\usepackage[numbers]{natbib}
\usepackage{algorithmicx}
\usepackage[ruled]{algorithm2e}
\usepackage{algpseudocode}
\usepackage[utf8]{inputenc} 
\usepackage[T1]{fontenc}    
\usepackage{hyperref}       
\usepackage{url}            
\usepackage{booktabs}       
\usepackage{amsfonts}       
\usepackage{nicefrac}       
\usepackage{microtype}      
\usepackage{lipsum}
\usepackage{fancyhdr}       
\usepackage{graphicx}       
\graphicspath{{media/}}     

\title{DeepGrav: Anomalous Gravitational-Wave Detection Through Deep Latent Features
}

\author{
  Jianqi Yan$^{\ast}$, Alex P. Leung$^{\dagger}$ \\
  Department of Physics \\
  The University of Hong Kong \\
  Hong Kong\\
  \texttt{\{easonyan, alexpl\}@hku.hk} \\
   \And
 Zhiyuan Pei $^{\ast}$\\
 Faculty of Innovation Engineering \\
 Macau University of Science and Technology \\
 Macau \\
 \texttt{2109853eii30002@student.must.edu.mo} \\
   \And
  David C. Y. Hui$^{\dagger}$, Sangin Kim \\
  Department of Astronomy and Space Science,\\ 
  Chungnam National University,\\ 
  Daejeon 34134, Republic of Korea\\
  \texttt{huichungyue@gmail.com}\\ 
  \texttt{kimsanginn@gmail.com}\\
}

\begin{document}
\maketitle

\begin{abstract}
This work introduces a novel deep learning-based approach for gravitational wave anomaly detection, aiming to overcome the limitations of traditional matched filtering techniques in identifying unknown waveform gravitational wave signals.
We introduce a modified convolutional neural network architecture inspired by ResNet that leverages residual blocks to extract high-dimensional features, effectively capturing subtle differences between background noise and gravitational wave signals.
This network architecture learns a high-dimensional projection while preserving discrepancies with the original input, facilitating precise identification of gravitational wave signals.
In our experiments, we implement an innovative data augmentation strategy that generates new data by computing the arithmetic mean of multiple signal samples while retaining the key features of the original signals.\\

In the {\it NSF HDR A3D3: Detecting Anomalous Gravitational Wave Signals} competition, it is honorable for us (group name: easonyan123)  to get to the first place at the end with our model achieving a true negative rate (TNR) of 0.9708 during development/validation phase and 0.9832 on an unseen challenge dataset during final/testing phase, the highest among all competitors.
These results demonstrate that our method not only achieves excellent generalization performance but also maintains robust adaptability in addressing the complex uncertainties inherent in gravitational wave anomaly detection.
\end{abstract}

\keywords{Gravitational Waves  \and Anomaly Detection \and Deep Learning \and HDR A3D3 Competition}

\renewcommand{\thefootnote}{\fnsymbol{footnote}}
\footnotetext{$^\ast$ These authors contributed equally to this work.}
\footnotetext{$^\dagger$ Corresponding author}
\renewcommand{\thefootnote}{\arabic{footnote}}

\section{Introduction}
The detection of gravitational waves have led to a paradigm shift in the way we explore our Universe. They allow us to study the extraordinary events such as the merging of two black holes directly \citep[e.g.][]{PhysRevLett.116.061102} and test the gravitational theories at strong field limit \citep[e.g.][]{PhysRevD.103.024041}. Despite a large number of gravitational wave events have been found, their diversity is rather limited. All these events belong to the class of compact binary coalescence (CBC) which originates from the mergers of black hole binaries, neutron star binaries, or the systems comprise a black hole and a neutron star \footnote{see \url{https://gwosc.org/eventapi/html/allevents/}}. 

The conventional method for searching CBC events is matched filtering \citep[cf.][]{Abbott_2020}, which is done by cross-correlating a template of known waveform and the gravitational wave data, which is in the form of time series, at different time delays to produce a filtered output. It can be proved that the signal-to-noise ratio, which is defined as the ratio of the value of the filtered output to the root mean square value for the noise, can be optimized by a matched filter comprises the ratio of the template of the actual waveform to the spectral noise density of the interferometer \cite{PhysRevD.95.042001}. 

The success of matched filtering in revealing many CBC events thanks to our knowledge of their waveform. However, such technique cannot be applied to the cases that their waveforms are poorly modeled or even unknown. Ascribing to this limitation, a lot of promising sources for gravitational waves (e.g. core-collapsed supernova, fast radio burst) have not yet been detected. Hence, the development of alternative detection methods is very important for further advancing gravitational wave astrophysics.

\section{The Competition}
\subsection{Dataset}
\label{sec:dataset}
The competition, {\it NSF HDR A3D3: Detecting Anomalous Gravitational Wave Signals}\footnote{\url{https://www.codabench.org/competitions/2626/}}, is a challenge for identifying astrophysical-like signals from the data prepared by the organizers \citep{campolongo2025buildingmachinelearningchallenges}. They prepared the 4 s baseline time series from the data collected by both Laser Interferometer Gravitational-wave Observatory (LIGO) detectors (Hanford and Livingston) during the first half of the third operation run (O3a) with a sampling rate of 4096 Hz. All these baselines have  transient instrumental artifacts (glitches) and known gravitational wave events removed. These events/glitch-free baselines are served as the background for injection of simulated signals. 

Two classes of the simulated signals were injected for the competition: 1. Binary Black Hole (BBH); 2. Sine-Gaussian Low-Frequency (SGLF). While BBH signals mimicking characteristic CBC events, SGLF are simply ad-hoc model as a proxy for the gravitational wave signal with unclear origin. The locations of both signal classes were uniformly distributed over the sky to facilitate the projection of simulated signals into the waveforms in both LIGO detectors. Apart from BBH and SGLF, there are also samples that no signal was injected which are taken as the class of background. 

They have further pre-processed the data by spectral whitening and band-pass filtering for a range of 30-1500 Hz. For avoiding the edge-effects from pre-processing, 1 s intervals were removed from each end. 
For each sample, 200 data points (corresponding to 50 ms at a sampling rate of 4096 Hz) were extracted from the remaining 2 seconds of data to ensure an appropriate length of data for training.
Eventually, three classes of data were provided in total: Background, BBH, and SGLF, with each class containing a tensor of dimensions $(100000,~200,~2)$ that represents the number of samples, the number of sequences, and the number of detectors, respectively. 

\subsection{Implementation Ideas}
\label{sec.ideas}

\begin{figure}
    \centering
    \includegraphics[scale=0.5]{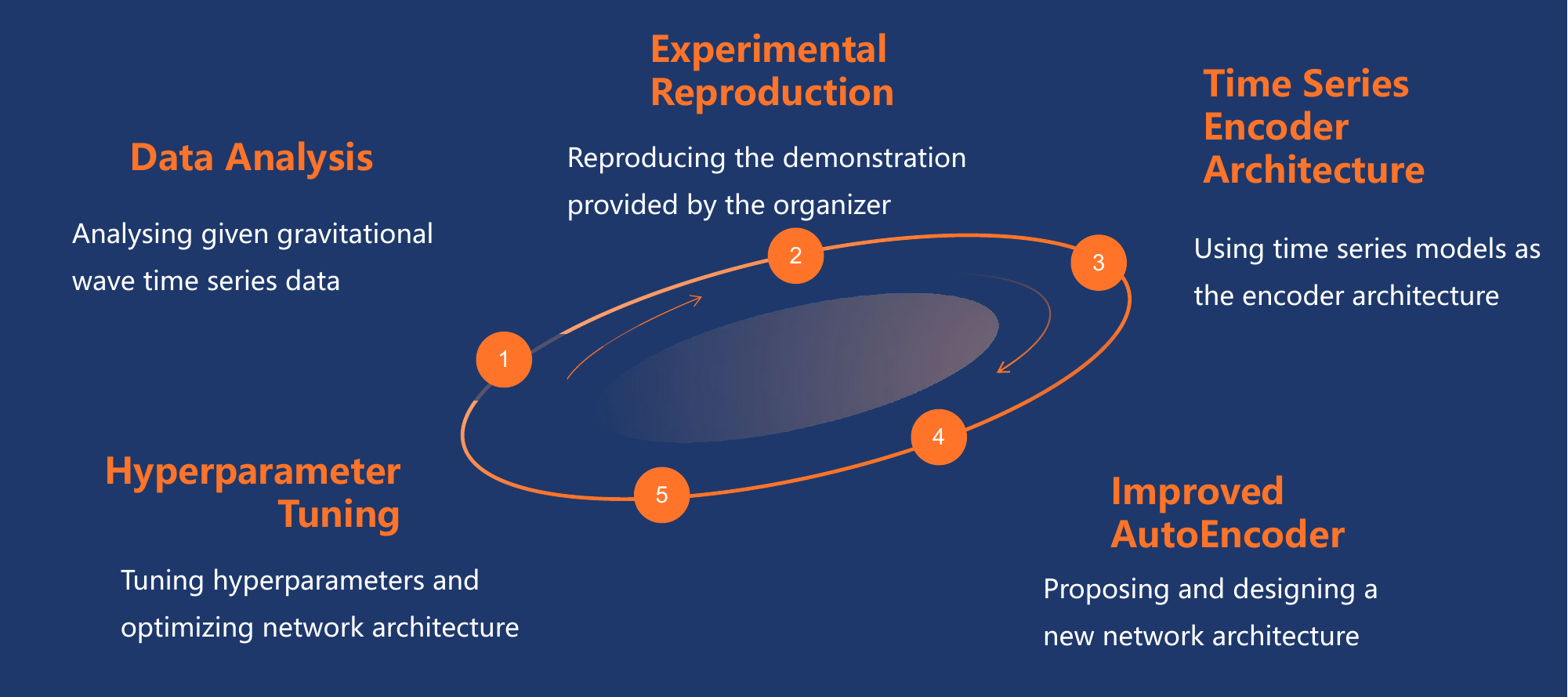}
    \caption{Illustration of the development process and key steps in the gravitational wave anomaly detection competition.}
    \label{fig.idea}
\end{figure}

In the competition, we first analyze the time series data provided by the organizers.
There are three subsets are provided, including background signals, simulated Binary Black Hole (BBH) mergers signal and Sine-Gaussian Low Frequency (SGLF) signals.
We then study the organizers’ sample code, which employs an AutoEncoder \citep{bank2023autoencoders} architecture that integrates a Transformer \citep{vaswani2017attention} encoder with a dense \citep{huang2017densely} decoder to differentiate between background signals and gravitational wave signals.
The key idea of using AutoEncoder in anomaly detection is that using self-supervised learning to reconstruct background signal so that the presence of a gravitational wave signal leads to a significant increase in reconstruction error.

We then run the codes and reproduce the experiments to evaluate the performance of using AutoEncoder with Transformer encoder.
Our results indicate that the Autoencoder architecture is not effective for this task.
Since the Autoencoder is designed to produce outputs that mirror its inputs, it becomes highly dependent on the training data and consequently exhibits high reconstruction errors for unseen background samples.
Moreover, the experimental outcomes indicate that no clear hyperplane exists to separate BBH or SGLF signals from the background, highlighting the similarity in the data representations.

Subsequently, we attempt to replace the encoder with other advanced time series architectures, such as TsMixer \citep{ekambaram2023tsmixer} and Mamba \citep{gu2023mamba}.
However, these modifications still did not yield the expected improvements.
This suggests that the limitation is not due to the Transformer’s capacity to represent the input data but rather that the Autoencoder architecture is not well-suited for this task. 
Finally, we propose an improved network architecture tailored to better address this problem, with detailed design specifications provided in Section \ref{sec:architecture}.

\section{Our proposed Network Architecture}
\label{sec:architecture}

In this section, we mainly describe our improved network architecture.
We begin by inputting $200\times2$ time series data into a convolutional network inspired by ResNet\citep{he2016deep}.
High-dimensional features are extracted by multiple convolutional layers, batch normalization \citep{ioffe2015batch}, and max pooling, with residual blocks repeatedly incorporated into the network.
Within each residual block, the network learns a high-dimensional projection output $\mathbf{C}$ and explicitly retains the difference between $\mathbf{C}$ and the original input ($\mathbf{C} - input$).
This design enables the model to capture the discrepancies in high-dimensional space when learning both background signals and various signals containing gravitational wave information.

\begin{figure}
    \centering
    \includegraphics[width=\linewidth]{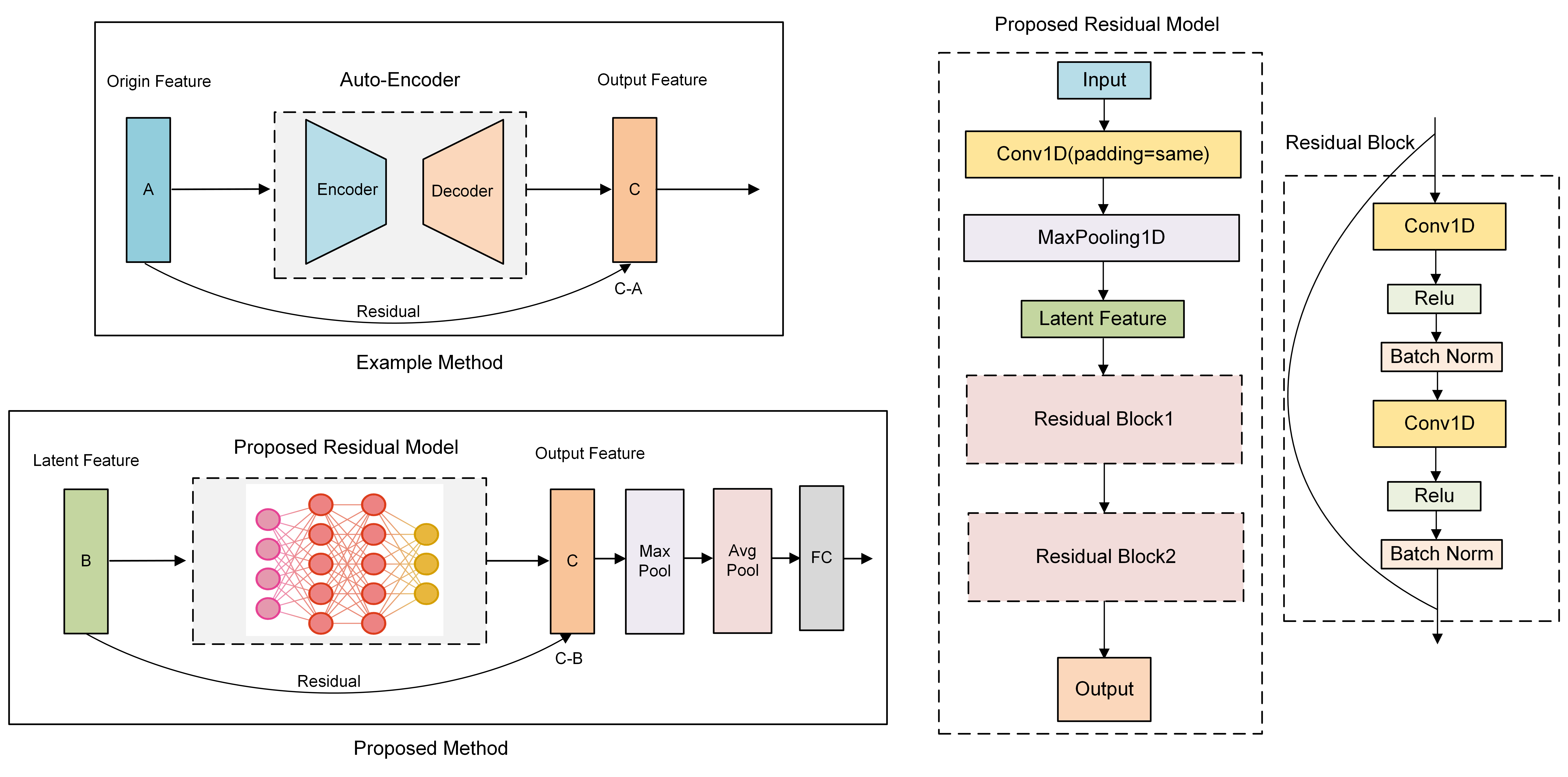}
    \caption{Illustration of our proposed network architecture.}
    \label{fig.na}
\end{figure}

The overall network architecture is illustrated in Figure \ref{fig.na}.
Furthermore, we introduce an improved residual model that obtains a high-dimensional feature representation $\mathbf{C}$ from the original input $\mathbf{A}$, followed by post-processing steps (max pooling, average pooling, and a fully connected layer) to further characterize the differential features between $\mathbf{A}$ and $\mathbf{C}$.

\section{Experimental Details}
\label{sec:exp}
\subsection{Experimental Setting}

In this experiment, the task is formulated as a binary classification problem.
The BBH and SGLF classes are combined into one category to facilitate their distinction from background signals.
The dataset is divided into three subsets: 70\% for training, 10\% for validation, and 20\% for testing.
The loss function adopted in this study is the cross-entropy loss for binary classification.
A Dense layer with two neurons combined with a Softmax activation function is applied at the output to perform binary classification, with label 0 corresponding to background data and label 1 to data containing gravitational-wave signals.
This design enables direct optimization using integer-based labels by minimizing the loss, effectively distinguishing between background and gravitational wave signals.

The training uses the NAdam \citep{dozat2016incorporating} optimizer with a learning rate of 0.0001, incorporating an early stopping mechanism that monitors the validation loss and terminates training if no improvement is observed over 10 epochs.
Additionally, an adaptive learning rate strategy reduces the learning rate to 10\% of its current value if the validation loss fails to decrease for 5 epochs.
The training is conducted for up to 100 epochs with a batch size of 512, and the dataset is randomly shuffled at the start of each epoch to improve generalization.
Upon completion of training, the model's performance in detecting gravitational-wave signals is evaluated by visualizing training and validation accuracy and loss trends, as well as assessing metrics on the validation set, including the Receiver Operating Characteristic (ROC) curves and AUC.
All our models and experimental settings are available on Github\footnote{\url{https://github.com/yan123yan/HDR-anomaly-challenge-submission}}.

\subsection{Data Augmentation}

In this experiment, we apply data augmentation separately to three different sets of training data.
We found that, for any signal, computing the arithmetic mean over $n$ instances generates new data while preserving essential information from the original signal.

\begin{equation}
    S' = \frac{\sum_{i=1}^n S_{i}}{n}.
   \label{eq:avg}
\end{equation}
where $S'$ denotes the generated signal and $S$ represents as the original signal.
For example, averaging two background signals produces a new background sample, whereas averaging two gravitational-wave signals retains critical features that distinguish them from background.
This concept constitutes the core of our data augmentation approach.

Moreover, by performing random sampling with replacement from the entire dataset and varying the number of signals combined, we can continuously create new data, as illustrated in Algorithm \ref{algo:aug}.
In this experiment, $n$ is set to 3, 5, and 10 and $N$ is set to 200,000 for each signal.
Finally, all augmented data are merged to form a new dataset.
Overall, there are 300,000 signals for each type are utilized to train the model by merging augmented data.
During the model tuning phase, this data augmentation method improves the accuracy from 98.8\% to 99.5\% on the test set, indicating that this 
approach is both efficient and effective for anomaly detection in gravitational-wave signals.

\begin{algorithm}[htbp]
	\label{algo:aug}
	\caption{Data augmentation for background, BBH, SGLF Signals}
	\KwIn{Number of augmented signals $N$, number of selected signals $n$, and selected signals from the training set $\mathbf{S}$.}
	\KwOut{Augmented signals $\mathbf{S'}$.}
	\BlankLine
	Initialize the augmented signal set: $\mathbf{S'} \gets \emptyset$\;
	\For{$i=1$ \KwTo $N$}{
	    Randomly select $n$ signals from $\mathbf{S}$ to form the subset $\mathbf{S}_{\text{sub}}$\;
	    Compute the average:
	    \[
	    \mathbf{S'}_i \gets \frac{1}{n} \sum_{j=1}^{n} \bigl({\mathbf{S}_{\text{sub}}}\bigr)_j\;
	    \]
	}
\end{algorithm}

\subsection{Experimental Results}
In this competition, 272 teams participated, generating a total of 2478 submissions.\footnote{\url{https://www.codabench.org/competitions/2626/}}
The event was structured into two phases: a development/validation phase, during which each team could submit up to 500 entries, and a final/testing phase, which permitted only one model submission per team.
To ensure a fair comparison, the organizers employed the True Negative Rate (TNR) as the evaluation metric, defined as:

\begin{equation}
   \text{TNR} = \frac{\text{TN}}{\text{FP} + \text{TN}}
   \label{eq:tnr}
\end{equation}
where TN (True Negatives) denotes the number of negative samples correctly identified, and FP (False Positives) refers to the number of negative samples misclassified as positive.

\begin{figure}
    \centering
    \includegraphics[scale=0.6]{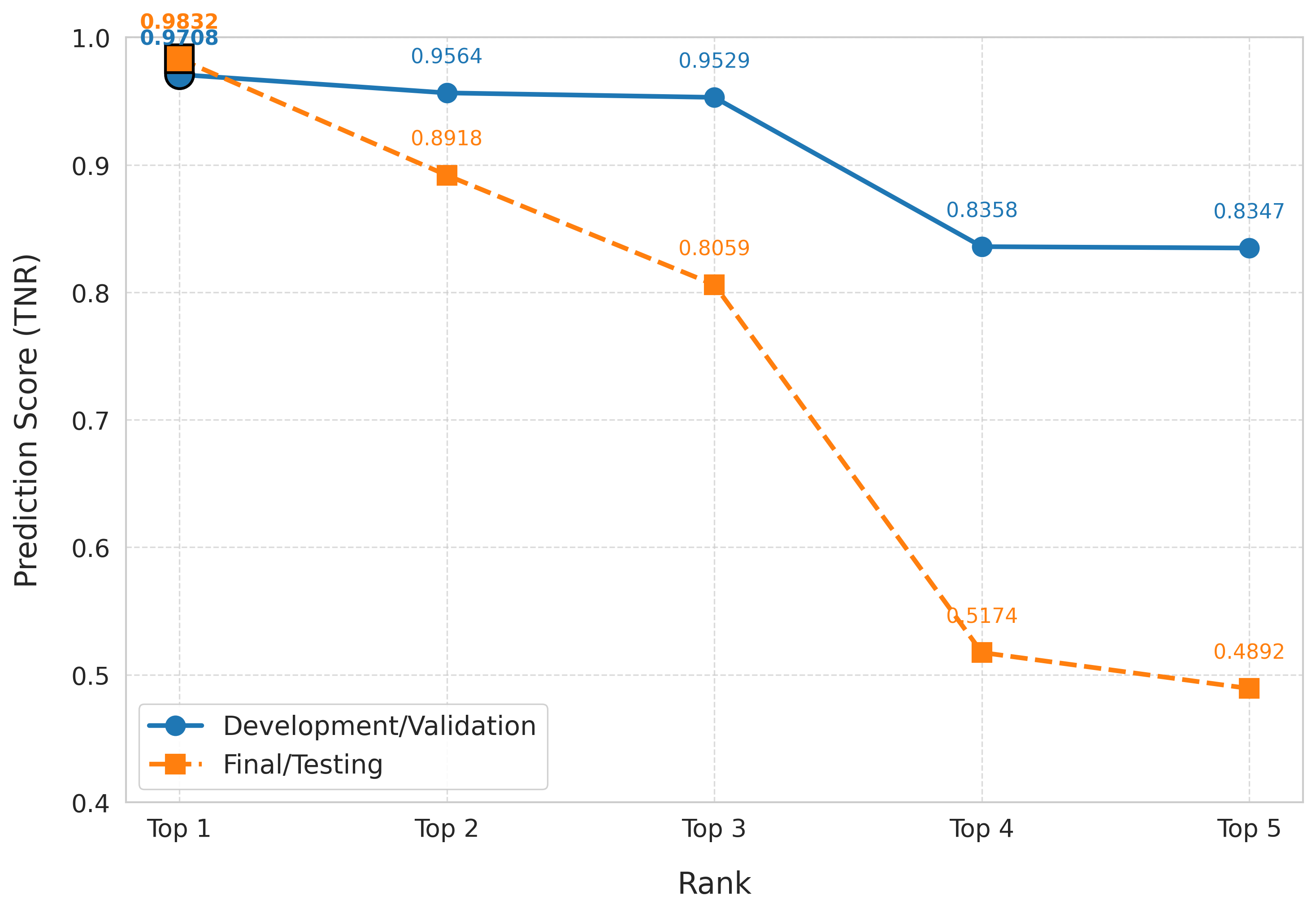}
    \caption{True negative rate performance comparison of top-ranked models in development/validation and final/testing phases of this competition.}
    \label{fig.result}
\end{figure}

Figure \ref{fig.result} compares the best performances of our team with those of other participants throughout both phases.
Our model achieves the highest TNR of 0.9708 in the development/validation phase and 0.9832 in the final/testing phase.
Notably, our approach also deliver superior results on an entirely unseen challenge dataset, underscoring the excellent generalization capability of our method.
This demonstrates that our model not only excels on known datasets but also exhibits robust adaptability and resilience in addressing the complex and uncertain challenges inherent in gravitational wave anomaly detection.


\section*{Acknowledgments}
D.C.Y.H. is supported by the research fund of Chungnam National University and by the National Research Foundation of Korea grant 2022R1F1A1073952.

\bibliographystyle{unsrt}  
\bibliography{references}

\end{document}